\documentclass[runningheads]{llncs}
\usepackage{graphicx}
\usepackage{tikz}
\usepackage{comment}
\usepackage{amsmath,amssymb} % define this before the line numbering.
\usepackage{color}

\usepackage[accsupp]{axessibility}  % Improves PDF readability for those with disabilities.

\usepackage{wrapfig}
\usepackage[ruled]{algorithm2e}
\usepackage{multirow}
\usepackage{hyperref}

\makeatletter
\newcommand{\thickhline}{%
    \noalign {\ifnum 0=`}\fi \hrule height 1pt
    \futurelet \reserved@a \@xhline
}
\makeatother

\begin{document}
% \linenumbers
\pagestyle{headings}
\mainmatter
\def\ECCVSubNumber{6024}  % Insert your submission number here

\title{
    Ensemble Knowledge Guided Sub-network Search\\ and Fine-tuning for Filter Pruning
    } % Replace with your title

\titlerunning{Ensemble Knowledge Guided Sub-network Search and Fine-tuning}
\author{
Seunghyun Lee \orcidID{0000-0001-7139-1764} \and
Byung Cheol Song\orcidID{0000-0001-8742-3433}
}
\authorrunning{S. Lee and B. C. Song}
\institute{Inha University, Incheon, Republic of Korea\\
\email{lsh910703@gmail.com, bcsong@inha.ac.kr}}

%******************
\maketitle

%%%%%%%%% ABSTRACT
\begin{abstract}
    Conventional NAS-based pruning algorithms aim to find the sub-network with the best validation performance.
    However, validation performance does not successfully represent test performance, i.e., potential performance.
    Also, although fine-tuning the pruned network to restore the performance drop is an inevitable process, few studies have handled this issue.
    This paper provides a novel Ensemble Knowledge Guidance (EKG) to solve both problems at once.
    First, we experimentally prove that the fluctuation of loss landscape can be an effective metric to evaluate the potential performance.
    In order to search a sub-network with the smoothest loss landscape at a low cost, we employ EKG as a search reward.
    EKG utilized for the following search iteration is composed of the ensemble knowledge of interim sub-networks, i.e., the by-products of the sub-network evaluation.
    Next, we reuse EKG to provide a gentle and informative guidance to the pruned network while fine-tuning the pruned network.
    Since EKG is implemented as a memory bank in both phases, it requires a negligible cost.
    For example, when pruning and training ResNet-50, just 315 GPU hours are required to remove around 45.04\% of FLOPS without any performance degradation, which can operate even on a low-spec workstation.
    the implemented code is available at \href{https://github.com/sseung0703/EKG}{here}.
L:/    
    \renewcommand{\thefootnote}{\fnsymbol{footnote}}
    \footnotetext{
    \textbf{Acknowledgments:}
    This work was supported by IITP grants funded by the Korea government (MSIT) (No.2021-0-02068, AI Innovation Hub and No. 2020-0-01389, Artificial Intelligence Convergence Research Center(Inha University)), and was supported by the NRF grant funded by the Korea government (MSIT) (No. 2022R1A2C2010095 and No. 2022R1A4A1033549).
    }
\end{abstract}

\section{Introduction}\label{sec1:intro}
    Network pruning is attracting a lot of attention as a lightweight technique to reduce computation and memory cost by directly removing parameters of deep neural networks (DNNs). In particular, filter pruning is advantageous in accelerating using the basic linear algebra subprograms (BLAS) library because it eliminates parameters in units of filters.
    Recently, filter pruning has been regarded as a kind of neural architecture search (NAS), that is, the sub-network search process. Actually, some methods using the existing NAS algorithms succeeded in finding a high-performance pruned network~\cite{tas,lu2020beyond,liu2019metapruning,manidp}. Based on `supernet' capable of dynamic inference, the other methods significantly reduced computational complexity while maintaining high performance of NAS-based pruning algorithm ~\cite{once-for-all,autoslim,dmcp,cafenet}.

    In general, NAS-based pruning algorithms consist of a search phase and a fine-tuning phase. We paid attention to the fact that prior arts seldom dealt with critical factors in each phase. First, note that rewards used in the search phase so far are not accurate enough to find the optimal sub-network. NAS-based pruning usually samples the validation set from the training dataset and uses the validation performance as a reward. In other words, it is assumed that validation performance has a high correlation with test performance, i.e., potential performance. 
    Although this approach improves the search speed greatly, it implies the possibility of overfitting the sub-network to the validation set. Despite such a risk, most previous works adopted validation performance as the reward without any doubt.
    \begin{wrapfigure}{T}{0.65\linewidth}\vspace{-0.8cm}
        \includegraphics[width=\linewidth]{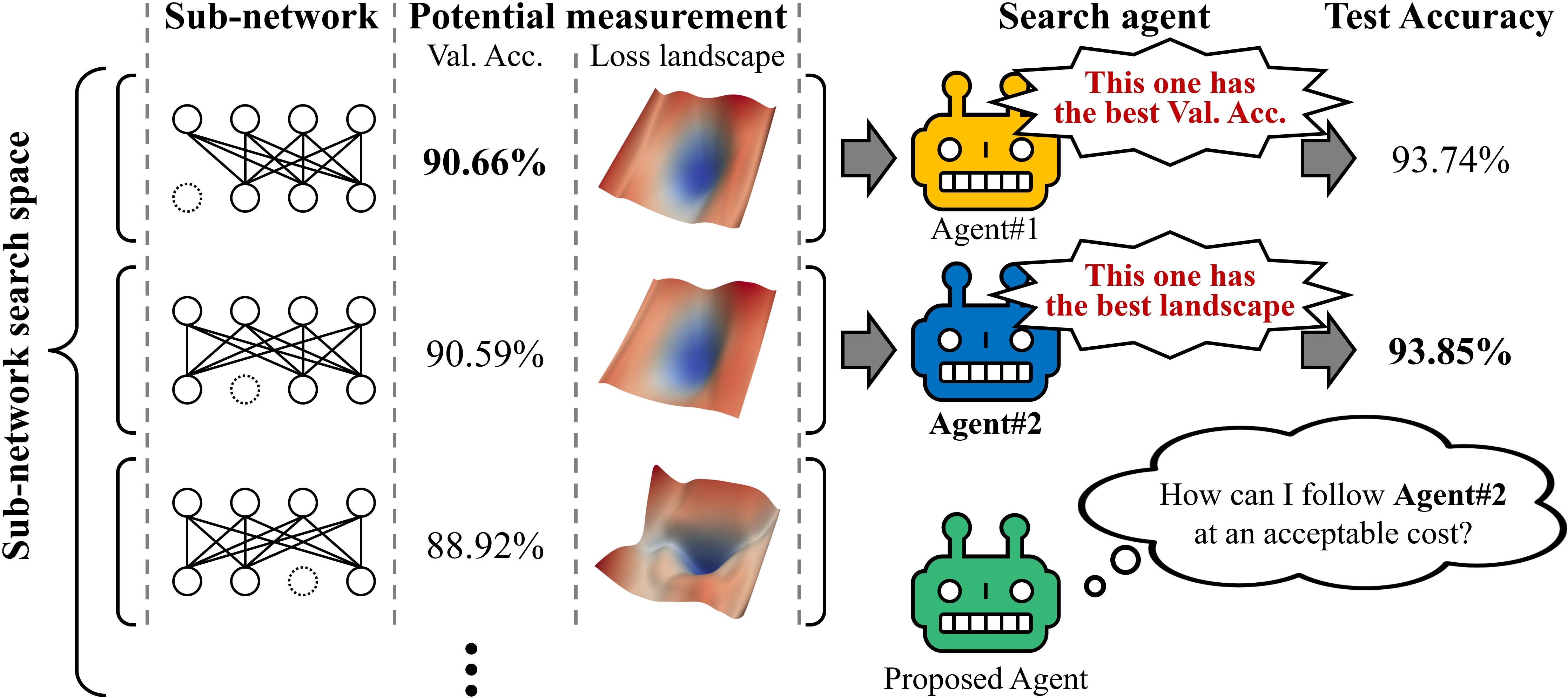}
        \vspace{-0.6cm}
        \caption{
            Conceptual visualization showing the performance change of the search algorithm according to reward and the intrinsic goal of the proposed agent
        }\label{fig1}
    \vspace{-0.6cm} \end{wrapfigure}
        
    Second, there were no prior arts considering the characteristics of the pruned network in the fine-tuning phase. During the filter pruning procedure, some information loss is unavoidable. To recover the information loss, we should fine-tune the pruned network. Since the pruned network still possesses the information of the pre-trained network, an approach differentiated from randomly-initialized networks is required. However, prior arts employed general learning strategies used when learning pre-trained networks as they are or adopted primitive knowledge distillation (KD).

    To solve two problems mentioned above, this paper presents a single solution, i.e., Ensemble Knowledge Guidance (EKG). In general, a student network constrained by teacher knowledge has a flatter local minima~\cite{ta,tan2018multilingual,chen2021robust} and faster convergence speed~\cite{tang2020understanding}, compared to ordinary optimization algorithms.
    Based on these characteristics, we formulate a way to improve the search and fine-tuning phases by using EKG as follows:
    First, we show that loss landscape fluctuation is a useful tool to evaluate the potential performance of sub-network. Fig.~\ref{fig1} visualizes the concept. The sub-network with the highest potential performance is expected to be sufficiently generalized as well as to have a fast learning speed. Since the smoothness of loss landscape can estimate both factors, i.e., generalization and learning speed~\cite{ll,bnwork,gotmare2018closer}, we employ it to evaluate sub-networks.
    However, since loss landscape-based reward requires massive computational cost, it is impractical to utilize the smoothness of loss landscape as a search reward. So, we use EKG to find an optimal sub-network with the smoothest loss landscape at an acceptable cost. Based on the property of KD, EKG implicitly selects a sub-network with a smoother loss landscape and discovers the optimal sub-network more precisely than previous works that solely use the validation loss.
    Here, as the source of knowledge, the output features of interim sub-networks, i.e., by-products of each search iteration, are stored and ensembled in the memory bank.
    Therefore, EKG incurs only a negligible cost because we don't have to infer the numerous genuine teacher networks in every training iteration.
    
    Furthermore, EKG is applied once again to fine-tune the pruned network.
    As mentioned above, KD is an effective way to improve a small network. However, as the pruning rate increases, the gap between the pre-trained and pruned networks also increases, which makes it difficult to transfer knowledge~\cite{ta}.
    To bridge the performance gap, we adopt interim sub-networks once again as teacher-assistant networks and build a memory bank with their knowledge.
    Then, according to the learning status of the pruned network, the knowledge of the memory bank is ensembled and transferred.
    Since the knowledge always maintains an appropriate gap with the pruned network, over-constraints can be avoided.
    
    As a result, the proposed EKG can improve the performance of the pruned network while affecting both phases with a marginal cost increase.
    For example, when pruning ResNet-50, 45.04\% FLOPS can be removed without any performance drop in only 315 GPU hours, which is an impressive result considering that our experiments are performed on a low spec workstation.
    Finally, the contributions of this paper are summarized as follows:
    \\$\bullet$ As a new tool to measure the potential performance of sub-network in NAS-based pruning, the smoothness of loss landscape is presented. Also, the experimental evidence that the loss landscape fluctuation has a higher correlation with the test performance than the validation performance is provided.
    \\$\bullet$ EKG is proposed to find a high potential sub-network and boost the fine-tuning process without complexity increase, which makes a high-performance light-weighted neural network democratized.
    \\$\bullet$ To our knowledge, this paper provides the world-first approach to store the information of the search phase in a memory bank and to reuse it in the fine-tuning phase of the pruned network. The proposed memory bank contributes to greatly improving the performance of the pruned network.
    
\section{Related Work}\label{sec2:rel}
    This section reviews the research trend of filter pruning. Existing filter pruning algorithms can be categorized into two groups. The first approach is filter importance scoring (FIS). FIS scores the importance of each filter according to a specific criterion and removes lower-score filters. Early FIS techniques adopted filter norms as criteria~\cite{norm-prune}. However, they faced with a problem of `smaller-norm-less-importance criteria.' So, various scoring techniques were proposed to solve this problem, e.g., gradient-norm~\cite{taylor,gbn}, geometric relation~\cite{fpgm,manidp}, and graph structure~\cite{lee_rilc,srr-gr}. Although the latest FIS techniques improved the performance of the pruned network, a question still remains as to whether the existing filter scores are closely related to the performance degradation.
    
    The second is a NAS-based approach and is more intuitive than the first. NAS-based methods define the number of filters as a search space and search for the optimal sub-network by learning and evaluating each sub-network~\cite{tas,lu2020beyond,liu2019metapruning}. Here, the performance itself is considered a score. However, the NAS-based methodology inevitably requires a huge cost because each sub-network must be actually trained. Thus, supernet was proposed to reduce this massive cost. A supernet in which a low index filter has higher importance than a high index filter is a network trained to enable dynamic inference~\cite{yu2018slimmable,dmcp,autoslim}. Once a supernet is trained, the search agent can explore the search space by free, so the search cost can be greatly reduced. Thus, supernet-based pruning has become the most dominant methodology. For example, \cite{cafenet,bcnet} achieved state-of-the-art (SOTA) performance by adopting a more sophisticated search algorithm, and \cite{yu2019universally,autoslim} presented a few effective methods for training supernets. However, since a supernet is not publicly available in general, it must be re-trained every time. In other words, supernet-based methods have a fatal drawback in that they cannot use well-trained networks~\cite{byol,yalniz2019billionscale}.
    
    The most important virtue of the pruning algorithm is to find a sub-network that minimizes information loss, but the process of restoring it is just as important. Surprisingly, how to fine-tune the pruned network has been rarely studied. For example, most pruning algorithms adopted a naive way of training the pruned network according to a general training strategy. If following such a naive way, a huge cost is required for fine-tuning. As another example, a method to re-initialize the pruned network according to lottery-ticket-hypothesis was proposed ~\cite{frankle2018lottery}. This method can guide the pruned network fallen into the local minima back to the global minima, but its learning cost is still huge. The last example is using KD~\cite{kd}. In general, the pre-trained network has a similar structure to the pruned network, but has a higher complexity and is more informative. So, we can intuitively configure the pre-trained and pruned networks as a teacher-student pair. Recently, a few methods have been proposed to improve the performance of the pruned network with the pre-trained network as a teacher~\cite{tas}. However, if the pruning rate increases, that is, if the performance gap between teacher and student increases, the information of the teacher network may not be transferred well~\cite{ta}. Also, since the pruned network has information about the target dataset to some extent, unlike general student networks, fine-tuning for the pruning algorithm is required.

\section{Method: Ensemble knowledge guidance}\label{sec3:method}
    \subsection{Accurate potential performance evaluation}\label{sec3.1:PPE}
        Conventional NAS-based pruning iteratively eliminates redundant filter set $\theta$ from the pre-trained network's filter set $\Theta_0$ until the target floating point operations (FLOPs) are reached. This process is expressed by
        \begin{equation}\label{eq1}
            \Theta_{i+1} = \Theta_{i} \setminus \theta^{*}_i
        \end{equation}
        \vspace{-0.6cm}
        \begin{equation}\label{eq2}
            \theta^{*}_i=\underset{\theta_i}{\text{argmax}}\ \mathcal{R}\left(\Theta_i, \theta_i \right)
        \end{equation}
        where $i$ is the search step, $\setminus$ is the set subtraction, and $\mathcal{R}$ indicates the reward of the search agent. It is costly to handle all filter combinations, so the agent usually divides the search space into intra-layer and inter-layer spaces. Then, the redundant filter set candidate $\phi_{i,l}$ is determined in each layer, and the candidate with the highest reward is selected. If a supernet is used as a pre-trained network, a candidate of each layer can be determined at no cost.
        As a result, the search process is regarded as a task that evaluates candidates, and is re-written by
        \begin{equation}\label{eq3}
            \Theta_{i+1} = \Theta_{i} \setminus \phi_{i,l^*}
        \end{equation}
        \vspace{-0.6cm}
        \begin{equation}\label{eq4}
            \begin{matrix}
                l^*=\underset{l}{\text{argmax}}\ \mathcal{R}(\Theta_i,\phi_{i,l})\\
                \text{s.t.}\ \Phi_i = \left\{\phi_{i,l} | 1\leq l \leq L\right\}
            \end{matrix}
        \end{equation}
        where $L$ stands for the number of layers in DNN.
        
        Among prior arts, some concentrated on how to construct candidates while maintaining the above-mentioned framework~\cite{yu2018slimmable,yu2019universally,autoslim,once-for-all}, and others presented new search agents to select the best candidate~\cite{bcnet,cafenet}. On the other hand, most of the previous studies employed sub-network's performance as a search reward $\mathcal{R}$ without any doubt. Since it is too heavy to use the entire training dataset for validation, the validation set $\mathcal{D}^{val}$ is sampled from the training dataset. Therefore, assuming that the loss function $\mathcal{L}$ is used as a metric to measure performance and $\phi_{i,l}$ is removed from $\Theta_i$ by a certain NAS-based pruning algorithm, the search reward is defined by
        \begin{equation}\label{eq5}
            \mathcal{R}(\Theta_i,\phi_{i,l}) = -\mathcal{L}(\mathcal{D}^{val};\Theta_i \setminus \phi_{i,l})
        \end{equation}
        \begin{figure}[t]\centering
            \includegraphics[width=\linewidth]{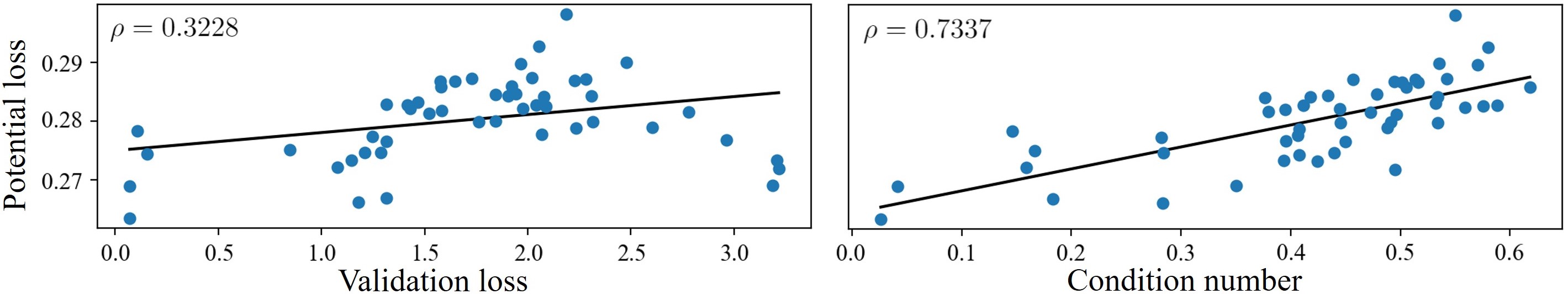}
            \vspace{-0.6cm}
            \caption{(Left) Potential loss vs. validation loss (Right) Potential loss vs. condition number. 50 sub-networks of ResNet-56 trained on CIFAR10 were used for this experiment}\label{fig2}
        \vspace{-0.5cm}\end{figure}
        \indent The reason to search for the sub-network with the highest validation performance is that its potential performance after fine-tuning is expected to be high. In other words, it is assumed that validation performance is highly correlated with potential performance. To examine whether this assumption holds, we randomly sampled sub-networks of ResNet-56~\cite{resnet} trained on CIFAR10~\cite{cifar} and measured the validation performance and test performance after fine-tuning. Specifically, Pearson Correlation Coefficient~\cite{pcc} (PCC) $\rho_{X,Y}$ of the validation loss (X) and potential loss (Y) was calculated. For the detailed training configuration of this experiment, please refer to the supplementary material. The left plot of Fig.~\ref{fig2} shows that the correlation between the two losses is not so high. Especially, note that the validation loss has high variance in the low potential loss region of the most interest. This phenomenon makes it difficult to adopt the validation loss as a reliable reward. Since the search process is a sort of optimization, exploring the validation set makes the sub-network fit the validation set. In other words, there is a risk of overfitting. Therefore, instead of simply measuring performance on a given dataset, we need another indicator to measure generality.

        We introduce loss landscape~\cite{ll} as a means to analyze the generality of sub-networks. The more the loss landscape is smooth and close to convex, the more the network is robust and easy to be trained. Therefore, quantitatively measuring fluctuations in the loss landscape can determine how well the network is generalized, which provides higher-order information than performance. Because a network with fewer filters has relatively large fluctuations in the loss landscape, the loss landscape of the sub-network becomes more complex as pruning progresses. However, if the information of a certain filter is redundant, the generality error as well as fluctuations in the loss landscape do not increase. Based on this insight, we assume that a sub-network can be evaluated through the fluctuation of the loss landscape. To verify whether this assumption is valid, we examine the PCC of loss landscape fluctuation (X) and potential loss (Y). As an index representing the loss landscape fluctuation, we employed the condition number (CN), i.e., the ratio of minimum and maximum eigenvalues of the Hessian~\cite{cn,ll}, which is defined by
        \begin{equation}\label{eq6}
            \text{Condition number} = \left| \lambda^{\text{min}}/\lambda^{\text{max}} \right|
        \end{equation}
        The right plot of Fig.~\ref{fig2} shows that CN has a higher correlation with potential loss than validation loss. In particular, note that the variance of CN is very small in the low potential loss region. This proves that CN is a more reliable indicator to evaluate sub-networks. Based on this experimental result, the next subsection designs a search process to select a sub-network with a smoother loss landscape.

    \subsection{Search with ensemble knowledge guidance}\label{sec3.2:EG}
        CN must be a better indicator to evaluate potential performance than validation performance. However, since the Hessian matrix required to calculate CN causes a huge amount of computation, it is burdensome to directly use CN as a search reward.
        
        To design a new sub-network search process of a reasonable cost, we adopt EKG. KD is widely used to enhance the performance of a small-size network (student) with a large-size network (teacher). Many researchers have pointed out that the performance of a student is improved because the loss landscape of the student gets smoother by receiving the teacher's knowledge, that is, the generality error of the student is reduced~\cite{ta,tan2018multilingual,chen2021robust}. Student networks are saturated with a smoother loss landscape under multi-directional constraints due to knowledge and target loss. However, if the pruning rate gets higher, a performance gap between the pre-trained network and the sub-network gets larger, making teacher knowledge not transferred effectively~\cite{ta}. Fortunately, interim sub-networks have suitable properties as teacher-assistant networks~\cite{ta}. In detail, interim sub-networks have intermediate performance between the pre-trained and pruned networks and are computationally efficient because there is no additional training cost.
        So, we build a memory bank and ensemble the knowledge of interim sub-networks at every step to keep the knowledge guidance at the middle point of the pre-trained network and the current sub-network.
        Finally, $\mathcal{R}$ is re-defined by
        \begin{equation}\label{eq7}
            \mathcal{R}(\Theta_i,\phi_{i,l}) = - \mathcal{L}(\mathcal{D}^{val};\Theta_i\setminus \phi_{i,l}) - \mathcal{L}(\mathcal{T}_i;\Theta_i\setminus \phi_{i,l})\\
        \end{equation}
        \vspace{-0.6cm}
        \begin{equation}\label{eq8}
            \mathcal{T}_i = \frac{1}{i+1}\sum_{j = 0}^{i} \mathcal{O}(\mathcal{D}^{val};\Theta_j)
        \end{equation}
        where $\mathcal{O}(\cdot;\cdot)$ indicates the inference result.
        \begin{figure}[t]\centering
            \includegraphics[width=0.6\linewidth]{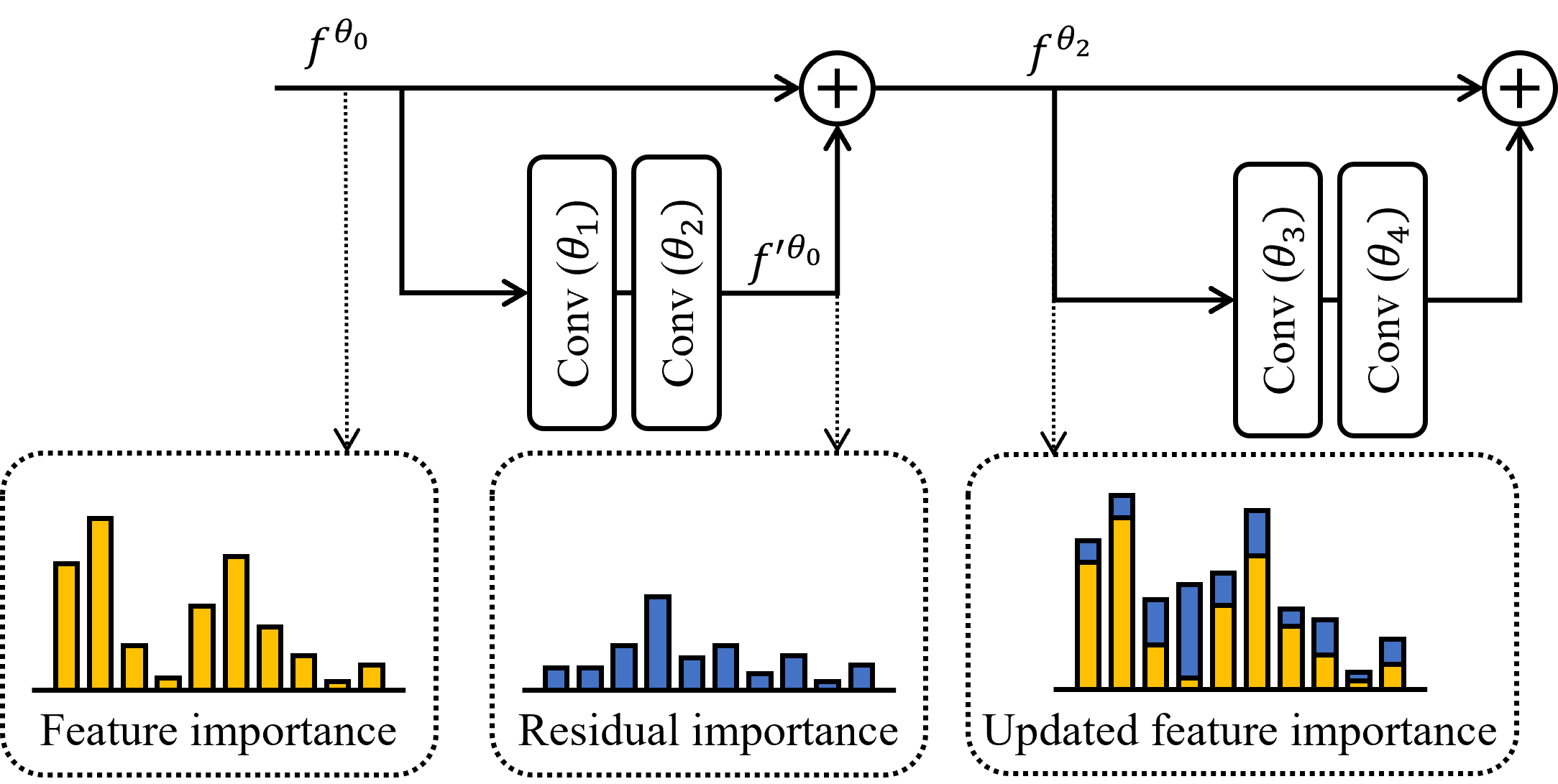}
            \caption{An example of a position sensing a feature map $f^{\theta_2}$ for scoring a filter $\theta_{2}$ in the proposed method. Compared with $f'^{\theta_2}$, which only grasps residual importance, $f^{\theta_2}$ can observe that importance is well measured regardless of architecture characteristics
            }\label{fig3}
            \vspace{-0.5cm}
        \end{figure}
        
        Based on the proposed reward, we reconfigure the existing search process as follows. First, a score is given to each filter according to Eq.~(7), and filters of low scores are selected as candidates. Here, in order to more accurately evaluate the importance of $\theta$, the input feature map of the next layer $f^{\theta}$ is used as in Fig.~\ref{fig3}. Since $f^{\theta}$ has both the information generated from $\theta$ and the characteristics of the network architecture, it can more accurately represent the importance of each filter. Using the well-known Taylor expansion~\cite{taylor,gbn}, we propose the following scoring function:
        \begin{equation}\label{eq9}
            \mathcal{S}(\theta) = \left| \frac{\partial\mathcal{R}(\Theta_i,\cdot)}{\partial f^{\theta}} f^{\theta} \right|
        \end{equation}
        The detailed derivation process of the above formula is depicted in the supplementary material.
        Next, filters of lower scores as much as a specific ratio $r$ for each layer become candidates, and a candidate with the highest reward is selected through a greedy search. Finally, the optimal sub-network $\Theta^{*}$ is searched iteratively. 
        
        In fact, the concept of transferring knowledge of a large sub-network to a small sub-network is already widely used for learning supernets~\cite{yu2018slimmable,autoslim,dmcp,once-for-all,cafenet}. We can say that the proposed reward based on ensemble knowledge inherits this concept. Here, scoring based on Taylor expansion can compute the scores of all filters at once, so it not only requires a much lower cost than learning the supernet itself but also has the advantage of using well-trained networks as they are.
        
    \subsection{Fine-tuning with ensemble knowledge guidance}\label{sec3.4:FT}
    % \subsection{Efficient fine-tuning via inverse-tracking}\label{sec3.4:FT}
        Even though a sub-network with high potential is available, the performance of the sub-network can be maximized through fine-tuning. So, we once again utilize EKG. As mentioned in Sec.~\ref{sec3.2:EG}, the pruning process generates many qualified teacher networks, i.e., interim sub-networks.
        If this knowledge is transferred to the pruned network again, the effect of inverse-tracking the information loss of the pruning process can be accomplished. However, acquiring knowledge by inferencing of all teacher networks at every iteration requires a huge computational cost.
        To transfer teacher knowledge efficiently, we propose to sample interim sub-networks with a uniform performance gap and store their knowledge in a memory bank.
        This entire procedure is expressed by
        \begin{equation}\label{eq10}
            \mathcal{M} = \{\mathbf{M}_{k} = \mathcal{O}(\mathcal{D}^{train},\mathcal{T}_{k}) | 1 \leq k \leq K \}
        \end{equation}
        \vspace{-0.6cm}
        \begin{equation}\label{eq11}
            \mathcal{T}_{k} = \underset{\Theta_i}{\text{argmin}}\left|\frac{K-k}{K}\mathcal{L}(\Theta^{*})
        + \frac{k}{K} \mathcal{L}(\Theta_0) - \mathcal{L}(\Theta_i)
        \right|
        \end{equation}
        where $K$ indicates the number of interim sub-networks to be sampled.
        
        Memory bank knowledge smoothly bridges the performance gap between pre-trained and pruned networks, but fixed knowledge can often cause overfitting. To resolve this side-effect, we employ contrastive learning. Contrastive learning is to minimize the gap between two data representations to which different augmentations are applied~\cite{simclr,byol}. So, it allows DNNs to learn more generalized representations. We set memory bank knowledge as the center of the augmented data distribution and transfer it into the sub-network. As in the search process, we select the teacher network from $\mathcal{M}$ according to the performance of the pruned network, and then the ensemble knowledge is transferred. Therefore, the loss function $\mathcal{L}^{ft}$ for fine-tuning is defined by
        \begin{equation}\label{eq12}
            \mathcal{L}^{ft}= \sum_{a=1}^{2} \mathcal{L}(\mathcal{D}^{train},\mathcal{A}_a;\Theta^{*})
            + \mathcal{L}(\overline{\mathbf{M}};\Theta_i\setminus \phi_{i,l})
        \end{equation}
        \vspace{-0.3cm}
        \begin{equation}\label{eq13}
            \overline{\mathbf{M}}=\mathbb{E}\left\{ \mathbf{M}_k \left|\qquad
                \begin{aligned}
            & 1\leq k \leq K\\
            & \mathcal{L}(\mathbf{M}_k) \leq \mathcal{L}(\mathcal{D}^{train};\Theta^{*})
                \end{aligned}\right.\right\}
        \end{equation}
        where $\mathcal{A}$ stands for augmentation function. The proposed fine-tuning strategy injects extra information into the pruned network at almost no cost. Also, it not only improves the learning speed but also has a regularization effect so that the pruned network has higher performance.

        Algorithm~\ref{alg1} summarizes the proposed method. The proposed search process has the following differentiation points from the existing supernet-based search algorithm.
        \\$\bullet$ Since the degree of freedom using the pre-trained network is higher and the potential performance is evaluated more accurately, a better sub-network can be obtained.  
        \\$\bullet$ The potential of sub-network is maximized by transferring the information lost in the search phase back to the fine-tuning phase, resulting in high performance.
        \\$\bullet$ Since the proposed search process has almost no additional cost compared to prior arts, it can operate even on a low-performance computer system.
        
        \begin{algorithm}[t]\small
            \caption{\small The proposed pruning algorithm}\label{alg1}
            \textbf{Input} : $\Theta_0, \mathcal{D}^{train}, r$ \hspace{.2cm} \textbf{Output} : Fine-tuned network\\
            \hspace{.2cm}1: Sample $\mathcal{D}^{subset}$ and $\mathcal{D}^{val}$ in $\mathcal{D}^{train}$ and fine-tune $\Theta_0$ in an epoch on $\mathcal{D}^{subset}$\\
            % \hspace{.2cm}2: Fine-tune $\Theta_0$ in an epoch on $\mathcal{D}^{subset}$\\
            \hspace{.2cm}2: Store initial ensemble knowledge $\mathcal{T}_0$\\
            \hspace{.2cm}3: $\textbf{Repeat}$\\
            \hspace{.2cm}4: \hspace{.45cm} Compute filter importance scores $\mathcal{S}(\theta)$ by Eq.~(\ref{eq9})\\
            \hspace{.2cm}5: \hspace{.45cm} Sample candidates $\Phi_{i}$ in each layers with $r$.\\
            \hspace{.2cm}6: \hspace{.45cm} Select $\phi_{i,l}^*$ by Eq.~(\ref{eq4}) that maximizes Eq.~(\ref{eq7}).\\
            \hspace{.2cm}7: \hspace{.45cm} Get next pruned network $\Theta_{i+1}$ by Eq.~(\ref{eq3}).\\
            \hspace{.2cm}8: \hspace{.45cm} Update ensemble knowledge $\mathcal{T}_{i+1}$ by Eq.~(\ref{eq8})\\
            \hspace{.2cm}9: \hspace{.5cm} $i = i+1$\\
            10: \textbf{Until} FLOPs reduction rate reaches the goal\\
            11: $\Theta^{*}=\Theta_{i}$\\
            12: Build memory bank by Eq.~(\ref{eq10})\\
            13: \textbf{Repeat} \\
            14: \hspace{.45cm} Get training sample in $\mathcal{D}^{train}$ and apply two augmentation functions.\\
            15: \hspace{.45cm} Minimize loss function in Eq.~(\ref{eq12}).\\
            16: \textbf{Until} Training is done
        \end{algorithm}
        
\section{Experimental results}\label{sec4:experiment}
    This section experimentally proves the performance of EKG. First, we verify that CN can be a metric that evaluates sub-networks more accurately than validation performance. Second, an ablation study is given. Third, we prove that the proposed method accomplishes more effective pruning than conventional methods in various configurations and achieves SOTA performance at a reasonable cost. The datasets for experiments were CIFAR10, CIFAR100~\cite{cifar}, and ImageNet-2012~\cite{imagenet}. ResNet family~\cite{resnet} and MobileNet-v2~\cite{mobilenetv2} were used as network architectures. The proposed method was implemented in Tensorflow~\cite{tf}. The CIFAR pre-trained network was trained by ourselves, and the ImageNet pre-trained network is a version released in Tensorpack~\cite{tensorpack}. Detailed hyper-parameters for training are shown in the supplementary material. GPU hours were calculated on GTX1080Ti, and 1 to 2 GPUs were utilized, depending on the architecture.
    \begin{figure*}[t]\centering
        \includegraphics[width=\linewidth]{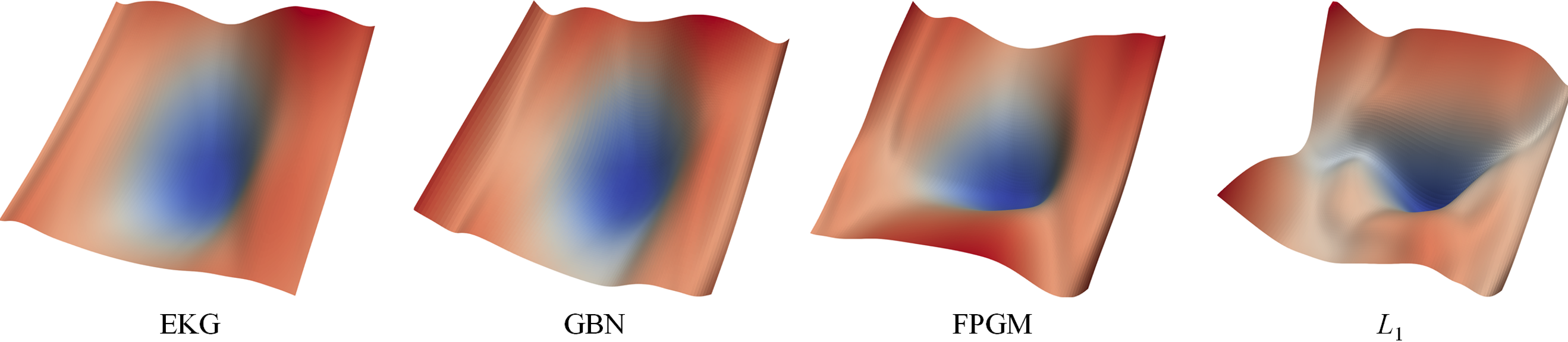}
        \caption{Visualization of loss landscapes of sub-networks searched by various filter importance scoring algorithms
        }\label{fig4}
    \end{figure*}
    \begin{table}[t]
        \centering\small
        \resizebox{0.7\linewidth}{!}{
        \begin{tabular}{cc|ccccc}
            \thickhline
            &Score       & $L_1$~\cite{norm-prune}   & FPGM~\cite{fpgm} & GBN~\cite{gbn} & EKG\\
            \thickhline
            \multirow{3}{*}{CIFAR10} & Validation loss  & 0.1569 & 0.1066 & \textbf{0.0710} & 0.0713\\
                                     & Condition number & 0.4078 & 0.1461 & 0.0414 & \textbf{0.0211}\\
                                     & Test accuracy    & 93.57  & 93.57 & 93.70 & \textbf{93.85}\\
            \hline
            \multirow{3}{*}{CIFAR100}&Validation loss   & 1.2050 & 0.7635 & \textbf{0.6816} & 0.7733\\
                                     &Condition number & 0.2535 & 0.0838 & 0.0747 & \textbf{0.0649}\\
                                     &Test accuracy    & 71.43 & 71.60   & 71.60 & \textbf{71.82}\\
            \thickhline
        \end{tabular}
        }
        \smallskip
        \caption{
        Comparison of validation loss, condition number, and test accuracy of various scoring methods on CIFAR10 and CIFAR100
        % Validation loss (Val. loss), condition number (CN), and test accuracy (Test Acc.) of various scoring methods.
        }\label{table1}
        \resizebox{\linewidth}{!}{
        \begin{tabular}{cc|ccc|ccc}
            \thickhline
            \multicolumn{2}{c|}{Teacher}&\multicolumn{3}{c|}{ResNet-56}&\multicolumn{3}{c}{MobileNet-v2}\\
            Search       & Fine-tune   & CIFAR10 & CIFAR100&GPU hours&CIFAR10&CIFAR100&GPU hours\\
            \thickhline
            \multicolumn{2}{c|}{Baseline}
                                & 93.84 & 72.62 & 0.44          & 94.21 & 76.07 & 1.83 \\
            \hline
            None & None         & 93.78 ($\pm$0.07) & 71.63 ($\pm$0.21) & 0.19\ /\ 0.50 & 93.69 ($\pm$0.06) & 74.27 ($\pm$0.18) & 0.31\ /\ 1.35   \\
            Single & None       & 93.54 ($\pm$0.09) & 71.66 ($\pm$0.17) & 0.21\ /\ 0.50 & 93.73 ($\pm$0.04) & 74.10 ($\pm$0.12) & 0.35\ /\ 1.35   \\
            Ensemble & None     & 93.85 ($\pm$0.10) & 71.82 ($\pm$0.20) & 0.21\ /\ 0.50 & 93.89 ($\pm$0.04) & 74.53 ($\pm$0.16) & 0.35\ /\ 1.35   \\
            Ensemble & Single   & 94.02 ($\pm$0.08) & 72.62 ($\pm$0.15) & 0.22\ /\ 0.88 & 94.44 ($\pm$0.09) & 76.11 ($\pm$0.15) & 0.38\ /\ 2.36   \\
            Ensemble & Ensemble & \textbf{94.09} ($\pm$0.07) & \textbf{72.93} ($\pm$0.16) & 0.22\ /\ 0.68 & \textbf{94.52} ($\pm$0.05) & \textbf{76.29} ($\pm$0.17) & 0.38\ /\ 1.64   \\
            \thickhline
        \end{tabular}
        }
        \smallskip
        \caption{Ablation study to verify each part of the proposed method. Here, None, Single, and Ensemble refer to three methods of using the teacher network. GPU hours were measured in search and fine-tuning phases, respectively}
        \label{table2}
        \vspace{-0.6cm}
    \end{table}

    \subsection{Empirical analysis for loss landscape-based evaluation}\label{sec4.1:EA}
        This subsection proves that the proposed method effectively searches the sub-network of smoother loss landscape. Unfortunately, direct comparison with existing supernet-based search algorithms is unreasonable. Instead, we assigned a score to each filter using a specific filter scoring technique ~\cite{norm-prune,fpgm,gbn}, and found a sub-network with the minimum validation loss and employed the result for comparison.

        First, we visualized the loss landscapes of the sub-networks for each method. As a network for visualization, ResNet-56 trained on CIFAR10 was used. Two directions for constructing the loss landscape were set to a gradient direction and its orthogonal random direction. Fig.~\ref{fig4} is the visualization result. We can observe that the sub-network of EKG has the least fluctuation. In particular, it is noteworthy that $L_1$-norm-based pruning, which has been adopted in many studies, is in fact not so effective. GBN~\cite{gbn} has the lowest fluctuation among conventional techniques.
        GBN has a similar structure to the proposed method because it is also based on Taylor expansion.
        However, since a reward of GBN relies on performance only and does not consider the information of the architecture sufficiently, its loss landscape fluctuation is larger than that of EKG.
        
        Next, let's compare the performance when sub-networks are trained with the same native fine-tuning (see Table~\ref{table1}). We could find that the correlation between the CN and the test accuracy is high. Even though EKG has a higher validation loss than GBN, its potential performance is expected to be higher because the CN or the loss landscape fluctuation of EKG is smaller.
        Actually, EKG's test accuracy on CIFAR10 was the highest as 93.85\%.
        
        Therefore, we can find that the proposed search process based on ensemble KD effectively finds the sub-network of smoother loss landscape. Also, the loss landscape fluctuation is a very accurate indicator for evaluating potential performance.

    \subsection{Ablation study}\label{sec4.2:AS}
        \begin{wraptable}{T}{0.5\linewidth}\centering\small
            \vspace{-0.8cm}
            \resizebox{0.8\linewidth}{!}{
            \begin{tabular}{c|ccc}
                \thickhline
                Method& Acc.  & FLOPs $\downarrow$ & Param $\downarrow$\\
                \thickhline
                TAS~\cite{tas}            & 93.69 & 52.7  & -    \\
                ABC~\cite{lin2020channel} & 93.23 & 54.13 & 52.20\\
                GAL~\cite{lin2019towards} & 91.58 & 60.2  & 65.9 \\
                GBN-60~\cite{gbn}         & 93.43 & 60.1  & 53.5 \\
                FPGM~\cite{fpgm}          & 93.49 & 52.6  & -    \\
                Hrank~\cite{Lin_2020_CVPR}& 93.17 & 50    & 57.6 \\
                LFPC~\cite{He_2020_CVPR}  & 93.34 & 52.9  & -    \\
                DSA~\cite{ning2020dsa}    & 92.93 & 52.6  & -    \\
                ManiDP~\cite{manidp}      & 93.64 & 62.4  & -    \\
                NPPM~\cite{nppm}          & 93.40 & 50.0  & -    \\
                SRR-GR~\cite{srr-gr}      & 93.75 & 53.8  & -    \\
                ResRep~\cite{resrep}      & 93.71 & 52.91 & -    \\
                GDP~\cite{gdp}            & 93.97 & 53.35 & -    \\
                \hline
                \multirow{2}{*}{EKG}     & 93.92 & 55.22 & 33.69\\
                                          & 93.69 & 65.11 & 46.52\\
                \thickhline
            \end{tabular}
            }
            \smallskip
            \caption{Performance comparison with several existing techniques. The baseline in this experiment is ResNet-56. Here, Acc is the accuracy, and FLOPs $\downarrow$ and Param $\downarrow$ are the reduction rates of FLOPs and the number of parameters, respectively}
            \label{table3}
            \vspace{-0.6cm}
        \end{wraptable}
        This section analyzes the effects of EKG on the search and fine-tuning phases.
        In this experiment, three scenarios of the teacher network are compared: None, Single (when only the pre-trained network is used), and Ensemble (when the proposed ensemble teacher is used). ResNet-56 and MobileNet-v2 trained with CIFAR10 and CIFAR100 were used as datasets, and the FLOPs reduction rate was set to 50\%. The experimental results are shown in Table~\ref{table2}.
        
        First, let's look at the sub-network search phase. Although the search process of `None' has almost no difference from that of GBN except for the score calculation position, `None' achieved some performance improvement over GBN.
        In the case of `Single,' over-constraint occurs due to the far performance gap between the pre-trained and interim sub-networks. Since interim sub-networks did not undergo fine-tuning to recover the information loss, this phenomenon is further highlighted.
        Thus, we can observe no performance improvement or rather a decrease in most configurations. For example, in the case of ResNet-56 on CIFAR10, the performance of `None' and `Single' was 93.78\% and 93.54\%, respectively. That is, using knowledge rather causes degradation of 0.24\%. On the other hand, if bridging the pre-trained and pruned networks through interim sub-networks, i.e., `Ensemble,' a sub-network of high potential performance can be selected because of more effective knowledge-based guidance. For example, in the case of ResNet-56 on CIFAR10, the performance of `Ensemble' is 0.31\% higher than `Single' and outperforms `None.'
        Here, it is noteworthy that the performance of `None' is already comparable than conventional algorithms (see Fig.~\ref{fig5}).
        Therefore, performance improvement of `Ensemble' is sufficiently acceptable considering that it requires almost no cost.

        Next, because the fine-tuning phase re-learns the sub-network, the over-constraints caused by the far performance gap are somewhat mitigated. Accordingly, even if only a single teacher is used, the performance of the pruned network is sufficiently improved, reaching 94.02\% in the case of ResNet-56 on CIFAR10. However, as the time for forwarding the teacher network is added, the training time increases by 71.09\% compared to `None.' On the other hand, when training with the proposed memory bank, which ensembles the knowledge of interim sub-networks, that is, in the case of `Ensemble' knowledge, we can achieve similar performance improvement with only 29.03\% increase in training time. This is because the memory bank provides qualified guidance by encapsulating information from many networks. Therefore, each part of the proposed method contributes to effectively providing a lightweight network with high performance at a low cost.
        
    \subsection{Performance evaluation}\label{sec4.3:eval}
        \begin{wrapfigure}{h}{0.6\linewidth} \vspace{-0.8cm} \centering
            \includegraphics[width=\linewidth]{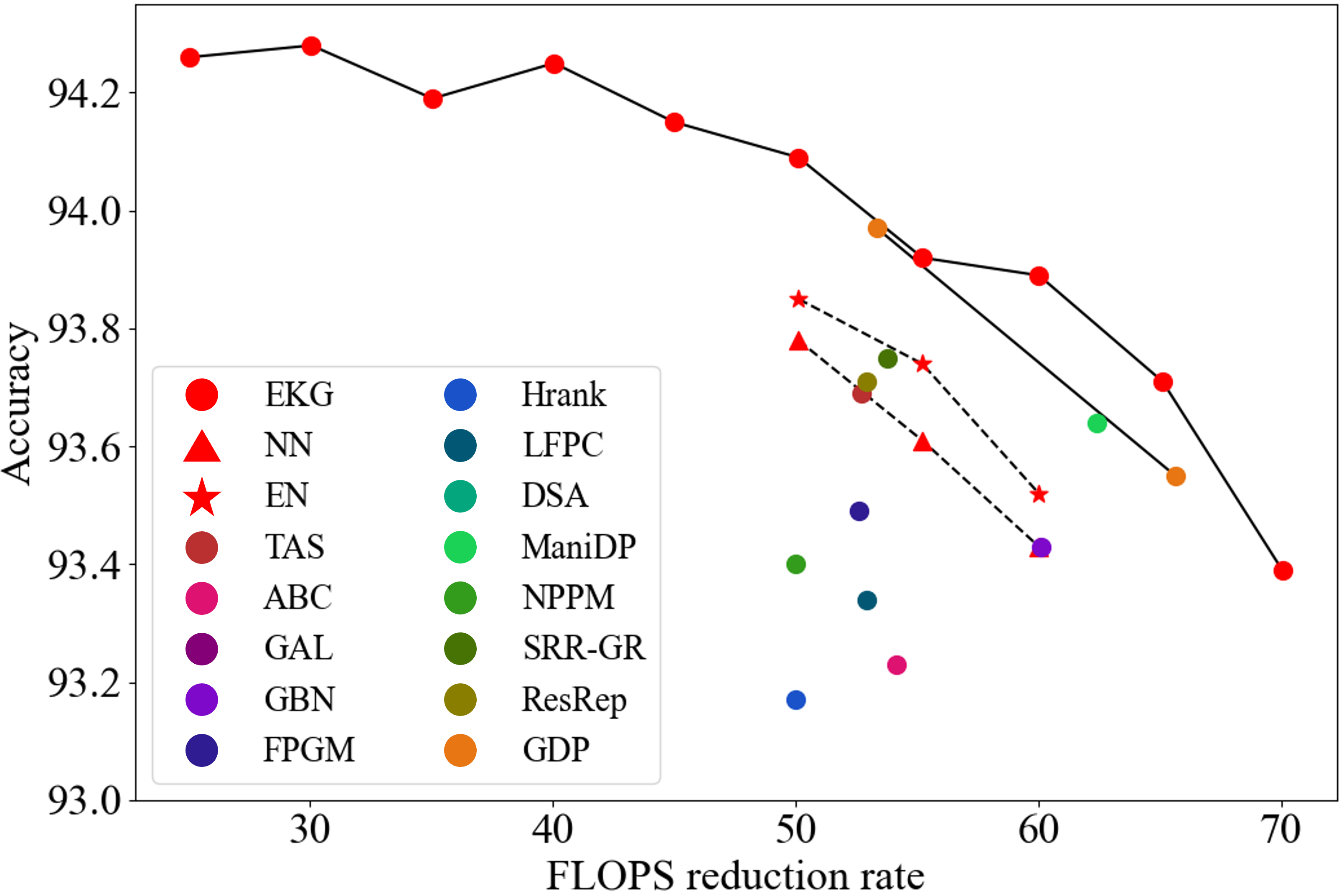}
            \vspace{-0.6cm}
            \caption{FLOPS reduction rate-accuracy of various pruning techniques for ResNet-56 on CIFAR10}\label{fig5}
        \vspace{-0.6cm}\end{wrapfigure}
        Experiments using various sizes of ResNet family~\cite{resnet} prove that the proposed method reaches SOTA performance. The datasets used in the experiments are CIFAR10 and ImageNet-2012.
        
        First, Table~\ref{table3} shows the experimental results for ResNet-56 trained on CIFAR10. For example, when 65.11\% of FLOPs are removed by the proposed method, the performance reaches 93.69\%. This proves that EKG outperforms the other methods with similar FLOPs reduction rates. To analyze the performance of EKG in detail, Fig.~\ref{fig5} plots the performances at various FLOPs reduction rates. We can observe that EKG provides higher performance than other methods at all FLOPs reduction rates.
        Note that NN (i.e, `None'), where any knowledge is not utilized in both phases, already shows a better or comparable performance than most conventional methods.
        Here, EN utilizing `Ensemble' knowledge only in the search phase gives sufficient performance improvement.
        This is because EKG accurately searches for a sub-network with high potential performance even at high reduction rates.
        In particular, the higher performance than the pre-trained network even at low FLOPs reduction rates shows that the proposed fine-tuning has a similar regularization effect even though the actual teacher network is not used.
        Also, this phenomenon indicates that better fine-tuning strategy is more crucial than better sub-network search.
        \begin{table}[t]
        \centering\small
            \resizebox{0.6\linewidth}{!}{
            \begin{tabular}{c|c|ccc}
                \thickhline
                ResNet & Method & Top-1 (diff.) & Top-5 (diff.)  & FLOPs $\downarrow$\\
                \thickhline
                \multirow{7}{*}{18}
                &TAS~\cite{tas}            & 69.15 (-1.50) & 89.19 (-0.68) & 33.3 \\ 
                &ABC~\cite{lin2020channel} & 67.80 (-1.86) & 88.00 (-1.08) & 46.9 \\ 
                &FPGM~\cite{fpgm}          & 68.41 (-1.87) & 88.48 (-1.15) & 41.8 \\
                &DSA~\cite{ning2020dsa}    & 68.62 (-1.11) & 88.25 (-0.82) & 40.0 \\
                &DMCP~\cite{dmcp}          & 69.20 & N/A & 43.0    \\
                &ManiDP~\cite{manidp}      & 68.88 (-0.88) & 88.76 (-0.32) & 51.0 \\
                &EKG                      & 69.39 (-0.99) & 88.65 (-0.87) & 50.1 \\
                \hline
                \multirow{5}{*}{34}
                &FPGM~\cite{lin2020channel} & 72.63 (-1.28) & 91.08 (-0.54) & 41.1 \\
                &SFP~\cite{fpgm}            & 71.84 (-2.09) & 89.70 (-1.92) & 41.1 \\
                &NPPM~\cite{nppm}           & 73.01 (-0.29) & 91.30 (-0.12) & 44.0 \\
                &ManiDP~\cite{manidp}       & 73.30 (-0.01) & 91.42 (-0.00) & 46.8 \\
                &EKG                       & 73.51 (-0.34) & 91.27 (-0.19) & 45.1 \\
                \hline
                \multirow{15}{*}{50}
                &DSA~\cite{dsa}            & 75.1  (-0.92) & 92.45 (-0.41) & 40.5 \\ 
                &FPGM~\cite{lin2020channel}& 75.59 (-0.56) & 92.63 (-0.24) & 42.7 \\
                &BNP~\cite{bnp}            & 75.51 (-1.01) & 92.43 (-0.66) & 45.6 \\
                &GBN~\cite{gbn}            & 76.19 (+0.31) & 92.83 (-0.16) & 41.0 \\ 
                &TAS~\cite{tas}            & 76.20 (-1.26) & 92.06 (-0.81) & 44.1 \\ 
                &SRR-GR~\cite{srr-gr}      & 75.76 (-0.37) & 92.67 (-0.19) & 45.3 \\
                &NPPM~\cite{nppm}          & 75.96 (-0.19) & 92.75 (-0.12) & 56.2 \\ 
                &ResRep~\cite{resrep}      & 76.15 (-0.00) & 92.89 (+0.02) & 54.9 \\
                &Autoslim~\cite{autoslim}  & 75.6  & N/A  & 51.6  \\
                &DMCP~\cite{dmcp}          & 76.20 & N/A  & 46.7  \\
                &CafeNet~\cite{cafenet}    & 76.90 & 93.3 & 52.0  \\
                &BCNet~\cite{bcnet}        & 76.90 & 93.3 & 52.0  \\
                &\multirow{2}{*}{EKG}     & 76.43 (-0.02) & 93.13 (-0.02) & 45.0 \\
                &                          & 75.93 (-0.52) & 92.82 (-0.33) & 55.0 \\
                &EKG-BYOL                 & 76.60 (-0.40) & 93.23 (-0.31) & 55.0  \\
                \thickhline
            \end{tabular}
            }
            \smallskip
            \caption{Comparison with various pruning techniques for ResNet family trained on ImageNet}
            \label{table4}
            \vspace{-0.6cm}
        \end{table}

        Next, Table~\ref{table4} shows the experimental results for ResNet family trained on ImageNet-2012. EKG achieved SOTA performance in all architectures. For instance, top-1 accuracy of EKG reached 73.51\% at 45.1\% FLOPs reduction rate for ResNet-34. Since EKG can be plugged-in easily, we adopted a primitive search algorithm, i.e., the greedy search of Autoslim~\cite{autoslim} to verify EKG. As a result, EKG improved Autoslim by 0.34\% in top-1 accuracy and by 3.5\% in FLOPs reduction rate for ResNet-50.
        On the other hand, in the case of ResNet-50, EKG showed lower performance than the latest supernet-based algorithms, i.e., CafeNet and BCNet. However, since EKG can employ well-trained networks which are not allowed for supernet-based algorithms, it can overcome even CafeNet and BCNet. For example, when using BYOL~\cite{byol}, which is the best among available Tensorflow-based ResNet-50, the top-1 accuracy of EKG was improved up to 76.6\% without cost increase. Therefore, if a better pre-trained network is released, EKG will be able to overwhelm the supernet in performance.
        
\begin{figure}[t]\centering
        \includegraphics[width=\linewidth]{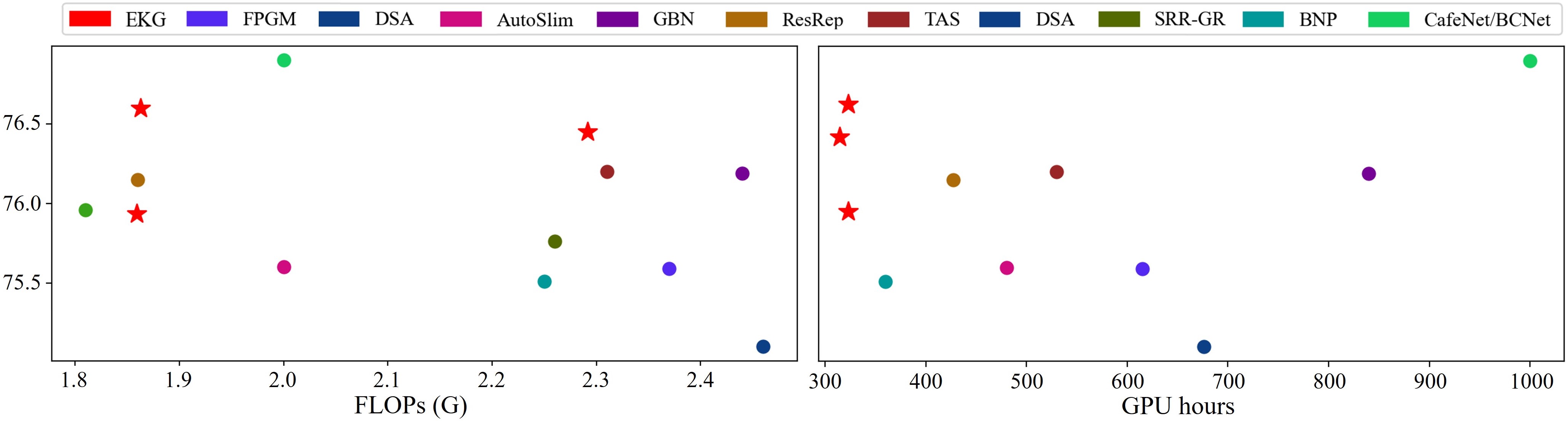}
        \caption{Performance analysis in case of ResNet-50 trained on ImageNet-2012. The left plot is the FLOPs reduction rate-Top-1 accuracy, and the right plot is the GPU hours-Top-1 accuracy
        }\label{fig6}
        \vspace{-0.3cm}
    \end{figure}

        For further analysis, the GPU hours-accuracy plot is given in Fig.~\ref{fig6}, together with the results for ResNet-50. The cost of pruning and fine-tuning ResNet-50 with EKG is about 315 GPU hours, which is much lower than the other methods with similar performance. In particular, EKG is computationally efficient because we used just mid-range GPUs mounted on general-purpose workstations. In other words, EKG democratizes DNNs because it allows more users to learn high-performance lightweight DNNs.
        
\section{Limitation}\label{sec5:limitation}
    EKG improves the performance of conventional NAS-based pruning by effectively searching and fine-tuning sub-networks. However, some limitations still remain. First, as mentioned above, CN looks impractical because it cannot directly replace performance due to computational complexity.
    In this paper, we indirectly induced the sub-network of the smoother loss landscape to be selected, but additional research is needed on how the CN can be used as a direct metric.
    Second, the fact that EKG has not been evaluated in sufficiently many configurations obscures its generalization a little bit.
    In order to analyze the performance of EKG more clearly, it is necessary to verify light-weight architectures such as MobileNet-v2 on a large dataset or to attach to other search algorithms, e.g., evolutionary search, NPPM, etc.
    
\section{Conclusion}\label{sec5:conclusion}
    This paper proposes EKG as an efficient and effective solution for search reward and fine-tuning strategy, which have been rarely considered in existing NAS-based pruning approaches. In particular, sub-network evaluation based on loss landscape fluctuations can reflect generality more accurately than validation performance which may cause a high risk of overfitting. Furthermore, EKG is pretty valuable because it can be easily plugged-in to the existing NAS-based pruning. Despite some unresolved limitations, we expect this paper to provide meaningful insight to filter pruning researchers. In particular, we pointed out and corrected the fact that inaccurate validation performance has been adopted without any doubt.

%%%%%%%%% REFERENCES
{
\bibliographystyle{splncs04}
\bibliography{egbib}
}
\section*{A.1 Details of the proposed filter importance scoring algorithm}\label{sec.A1}
    The proposed filter importance scoring (FIS) follows the general Taylor expansion-based methods~\cite{taylor,gbn}.
    Filter pruning aims to minimize $\Delta \mathcal{L}$, i.e., the difference in the target loss function when the pre-trained network $\Theta_0$ and a filter $\theta$ are removed.
    As described in this paper, to calculate the importance score of $\theta$, we use $f^{\theta}$, i.e., the input feature map of the next layer which is generated by $\theta$.
    For convenience in terms of notation, we redefine the target loss with $f^{\theta}$ removed as $\mathcal{L}(f^{\theta})$.
    Here, $\Delta \mathcal{L}$ is expressed as
    \begin{equation}\label{eqa1}
        \Delta \mathcal{L}(f^{\theta}) = \left | \mathcal{L}(f^{\theta}) -\mathcal{L}(0)  \right |
    \end{equation}
    Here, if $\mathcal{L}(0)$ is expanded using Taylor series, it is expressed as follows.
    \begin{equation}\label{eqa2}
        \begin{matrix}
            \mathcal{L}(0) = \sum_{n=0}^{\infty} \frac{ \mathcal{L}^{(n)}(f^{\theta}) }{n!} (0 - f^{\theta})^{n}\\\\
            \hspace{1.2cm} = \mathcal{L}(f^{\theta}) - f^{\theta} \nabla_{f^{\theta}} \mathcal{L}(f^{\theta}) + R_{1}
        \end{matrix}
    \end{equation}
    where $R_1$ is the Lagrange remainder. Next, Eq.~(\ref{eq1}) and Eq.~(\ref{eq2}) are combined. At this time, if $R_1$ is omitted to reduce the computational burden, it is expressed by the following formula.
    \begin{equation}\label{eqa3}
        \begin{matrix}
            \Delta \mathcal{L}(f^{\theta}) = \left | \mathcal{L}(f^{\theta}) - \mathcal{L}(f^{\theta}) + f^{\theta} \nabla_{f^{\theta}} \mathcal{L}(f^{\theta}) - R_{1}  \right |\\\\
            \approx \left| f^{\theta} \nabla_{f^{\theta}} \mathcal{L}(f^{\theta}) \right|\hspace{2.25cm}
        \end{matrix}
    \end{equation}
    The proposed method uses the reward $\mathcal{R}(\Theta_i,\cdot)$ instead of the target loss function.
    As a result, the proposed filter importance score $\mathcal{S}(\theta)$ for $\theta$ is expressed by
    \begin{equation}\label{eqa4}
        \mathcal{S}(\theta) = \left| \frac{\partial\mathcal{R}(\Theta_i,\cdot)}{\partial f^{\theta}} f^{\theta} \right|
    \end{equation}
    
    \begin{table}[t]
        \centering\small
        \begin{tabular}{c|ccc}
            \hline\hline
            Method& Acc.  & FLOPs $\downarrow$ & Param $\downarrow$\\
            \hline\hline
            \multirow{10}{*}{EKG}    & 94.26 & 25.06 &  8.84\\
                                      & 94.28 & 30.07 & 10.83\\
                                      & 94.19 & 35.05 & 16.25\\
                                      & 94.25 & 40.06 & 19.45\\
                                      & 94.15 & 45.00 & 22.62\\
                                      & 94.09 & 50.12 & 27.40\\
                                      & 93.92 & 55.11 & 33.69\\
                                      & 93.89 & 60.02 & 41.60\\
                                      & 93.69 & 65.11 & 46.52\\
                                      & 93.39 & 70.04 & 53.14\\
            \hline
        \end{tabular}
        \caption{Performance comparison with several existing techniques. The baseline in this experiment is ResNet-56. Here, Acc is the accuracy, and FLOPs $\downarrow$ and Param $\downarrow$ are the reduction rates of FLOPs and the number of parameters, respectively.}
        \label{tablea1}
    \end{table}

\section*{A.2 Experimental configurations}
    This section describes the configurations set up for various experiments in this paper.
    The kernels of convolutional layers and fully-connected (FC) layers were initialized by Xavier initializer~\cite{glorot2010understanding} and He initializer~\cite{resnet}, respectively.
    As hyper-parameters of batch normalization~\cite{batchnorm}, $\alpha$ and $\epsilon$ were set to 0.9 and $1\times10^{-5}$.
    The $L_2$-regularizer was applied to all trained parameters. Also, to train all networks, the stochastic gradient descent (SGD)~\cite{sgd} algorithm was used, and Nesterov accelerated gradient~\cite{nesterov} was applied. Here, momentum was set to 0.9.

    \subsection*{A.2.1 CIFAR}
        CIFAR10 and CIFAR100 use the same configuration as they have no difference except for the number of classes.
        
        \noindent\textbf{Pre-trained network:}
            ResNet-56~\cite{resnet} and MobileNet-v2~\cite{mobilenetv2} are learned for 200 epochs. Here, the batch size is 128. The initial learning rate is 0.1, which is reduced by 0.2 times at 60, 120, and 160 epochs, respectively. The weight of $L_2$-regularization is set to $5\times10^{-4}$ for ResNet-56 and $4\times10^{-5}$ for MobileNet-v2, respectively. Each data is normalized on a channel basis. Also, for data augmentation, random horizontal flipping and 32$\times$32 random cropping with padding of 4 pixel-thickness are used.

        \noindent\textbf{Sub-network search phase:}
            In the sub-network search phase, we divide the training dataset $\mathcal{D}^{train}$ into the training subset $\mathcal{D}^{subset}$ and the validation set $\mathcal{D}^{val}$. $\mathcal{D}^{subset}$ and $\mathcal{D}^{val}$ are configured by sampling 256 and 32 data per class so that there is no duplicate. Here, data augmentation is not used. In addition, batch normalization layers are set as training mode, and batch statistics are calculated and used each time. 
            
            Prior to the sub-network search phase, the pre-trained network is fine-tuned for 1 epoch with $\mathcal{X}^{subset}$. Here, the learning rate is set to $1\times10^{-3}$. The batch size for fine-tuning and evaluation is 256, and the ratio of the filters with low scores $r$ is 0.2.
            In the last iteration, $r$ is adjusted so that the FLOPs reduction rate is as close as possible to the given target pruning rate $\tau$.
        
        \noindent\textbf{Memory bank building phase:}
            The number of teachers $K$ to build a memory bank is 5. The settings of all other hyper-parameters are the same as those for evaluating in the sub-network search phase.
        
        \noindent\textbf{Fine-tuning by ensemble knowledge transfer:}
            ResNet-56 and MobileNet-v2 are trained only for 100 epochs, which is a half of a pre-trained network. Here, the batch size is 128. The initial learning rate is $1\times10^{-2}$, which decreases by 0.2 times at 30, 60, and 80 epochs, respectively. The weight of $L_2$-regularization is $5\times10^{-4}$. For additional data augmentation, random brightness, contrast, and saturation distortion are used. the temperature
    \begin{table*}[t]
        \centering
        \begin{tabular}{c|c|cccc}
            \hline\hline
            ResNet & Method & Top-1 (diff.) & Top-5 (diff.)  & FLOPs $\downarrow$ & GPU hour\\
            \hline\hline
            \multirow{15}{*}{50}
            &DSA~\cite{dsa}            & 75.1  (-0.92) & 92.45 (-0.41) & 40.5 & 675*\\
            &FPGM~\cite{lin2020channel}& 75.59 (-0.56) & 92.63 (-0.24) & 42.7 & 198*\\
            &BNP~\cite{bnp}            & 75.51 (-1.01) & 92.43 (-0.66) & 45.6 & 360\\
            &GBN~\cite{gbn}            & 76.19 (+0.31) & 92.83 (-0.16) & 41.0 & 530*\\ 
            &TAS~\cite{tas}            & 76.20 (-1.26) & 92.06 (-0.81) & 44.1 & 532*\\ 
            &SRR-GR~\cite{srr-gr}      & 75.76 (-0.37) & 92.67 (-0.19) & 45.3 & N/A\\
            &NPPM~\cite{nppm}          & 75.96 (-0.19) & 92.75 (-0.12) & 56.2 & N/A\\ 
            &ResRep~\cite{resrep}      & 76.15 (-0.00) & 92.89 (+0.02) & 54.9 & 427\\
            &\multirow{2}{*}{EKG}     & 76.43 (-0.02) & 93.13 (-0.02) & 45.0 & 315\\
            &                          & 75.93 (-0.52) & 92.82 (-0.33) & 55.0 & 310\\
            &EKG-BYOL                 & 76.60 (-0.40) & 93.23 (-0.31) & 55.0 & 315\\
            &Autoslim~\cite{autoslim}  & 75.6  & N/A  & 51.6  & 480\\
            &DMCP~\cite{dmcp}          & 76.20 & N/A  & 46.7  & N/A\\
            &CafeNet~\cite{cafenet}    & 76.90 & 93.3 & 52.0  & 1000*\\
            &BCNet~\cite{bcnet}        & 76.90 & 93.3 & 52.0  & 1000*\\
            \hline
        \end{tabular}
        \caption{Comparison with various pruning techniques for ResNet-50 trained on ImageNet. * indicates the estimated value.}
        \label{tablea2}
    \end{table*}
        
    \subsection*{A.2.2 ImageNet-2012}
        Since the basic training configuration of ImageNet-2012 is the same as that of CIFAR, this section describes only hyper-parameters.
    
        \noindent\textbf{Sub-network search phase:}
            $\mathcal{D}^{subset}$ and $\mathcal{D}^{val}$ are configured by sampling 256 and 32 data per class so that there is no duplicate. Prior to the sub-network search, the pre-trained network is fine-tuned for 1 epoch with $\mathcal{D}^{subset}$. Here, the learning rate is $1\times10^{-3}$. The batch size for fine-tuning and evaluation is 256, and the ratio of the filters with low scores $r$ is set to 0.1 for ResNet-18 and 0.2 for ResNet-34 and -50.
            
        \noindent\textbf{Memory bank building phase:}
            The number of teachers $K$ to build a memory bank is 5. The settings of all other hyper-parameters are the same as those for evaluating in the sub-network search phase.
        
        \noindent\textbf{Contrastive knowledge transfer:}
            ResNet-18, -34 -50 are trained for 100 epochs. Here, the batch size is 256. The initial learning rate is 0.1 and decreases by 0.1 times at 30, 60, and 90 epochs. The weight of $L_2$-regularization is set to $1\times10^{-4}$. We employ random brightness, contrast, saturation, and lighting distortion for additional data augmentation.

        \noindent\textbf{Estimation of GPU hours:}
            How to estimate GPU hours in Fig.~6 is described here. Most of previous works did not show such costs, but the approximate cost of some methods, i.e., FPGM~\cite{fpgm}, GBN~\cite{gbn}, DAS~\cite{ning2020dsa}, TAS~\cite{tas} can be estimated based on the released information. First, in the case of FPGM and GBN, how many epochs are required for pruning and fine-tuning are disclosed. In our environment, multiplying this by the GPU hours required to infer 1 epoch gives us the approximate GPU hours required by both techniques. For example, assuming epoch for forwarding is 0.5, GPU hours of FPGM and GBN are $100\times1.98=198$ and $((100000/1281167\times10\times0.5 + 10)\times20+60)\times1.98=530$, respectively. Also, DSA and TAS only disclosed GPU hours for ResNet-18. At this time, if it is assumed that computing time is proportional to FLOPs of ResNet-18 and ResNet-50, approximate GPU hours can be estimated. As a result, GPU hours of DSA and TAS are $300\times4.1/1.82=675$ and $236\times4.1/1.82=532$. 
            The authors mentioned in the paper that CafeNet~\cite{cafenet} and BCNet~\cite{bcnet} require a huge amount of computation to give up part of the algorithms when learning ImageNet. It is actually impossible to run the algorithms on a standard workstation. So, the costs of CafeNet and BCNet were expressed as `1000' symbolically in the paper.
            These estimates are not exact and may vary depending on the testing environment. However, it is useful enough as an index for estimating the approximate cost.
            
\section*{A.3 Experimental setting for random sub-networks}
    This section describes the configuration for the experiment in Figure 2 of this paper.
    First, the process of generating a random sub-network is as follows. A random sub-network is generated in a similar way to the proposed method's search process. At this time, in order to give randomness, selecting filter importance scores and candidates are randomly determined. In addition, in order to prevent the distribution of sub-networks intensively in low performance areas, we used the $L_{1}$-norm-based score for several trials so that a sub-network with higher performance is selected.
    
    Next, as mentioned above, when calculating the validation loss of the random sub-network, the loss is calculated by selecting the batch normalization layer as the training mode. In the case of the condition number, we analyze a 1-directional landscape composed by gradient direction.
    Here, if the loss landscape is traversed as much as the calculated gradient, the loss diverges, making it difficult to analyze the landscape. Thus, we traversed the 0.1 times boundary of gradients and analyzed the saturated part of the sub-network more intensively. The points where condition numbers were calculated were generated by uniformly sampling 21 points at the traverse boundary.
    And all the condition numbers are averaged and used as an index to evaluate the sub-network.
    
    Finally, the training of each random sub-network follows a plain fine-tuning strategy, and its training configuration is the same as that used when learning CIFAR mentioned above.
    
\section*{A.4 Numerical values for each plot result}
    Table~\ref{tablea1} and Table~\ref{tablea2} show the numerical value of each point at Figure 5 and the bottom of Figure 6 of this paper, respectively.
    
\end{document}